\documentclass{article}

\usepackage{arxiv}

\usepackage[utf8]{inputenc} 
\usepackage[T1]{fontenc}    
\usepackage{hyperref}       
\usepackage{url}            
\usepackage{booktabs}       
\usepackage{amsfonts}       
\usepackage{nicefrac}       
\usepackage{microtype}      
\usepackage{lipsum}		
\usepackage{graphicx}
\usepackage{amsmath}
\usepackage{enumitem}
\usepackage{doi}
\usepackage{multirow}

\usepackage{csquotes}
\usepackage[english]{babel}
\usepackage[doi=false,isbn=false,url=false,eprint=false]{biblatex}
\AtEveryBibitem{
    \clearfield{url}
    \clearfield{urlyear}
    \clearfield{urlmonth}
}

\title{Time Series Forecasting with Stacked Long Short-Term Memory Networks}

\date{} 					

\author{ 
	\href{}{\hspace{1mm}Frank Xiao}\thanks{Work is done during the internship at the \emph{Toronto Transit Commission}.} \\
	Operations Group\\
	Toronto Transit Commission\\
}

\begin{document}
\maketitle

\begin{abstract}
Long Short-Term Memory (LSTM) networks are often used to capture temporal dependency patterns. By stacking multi-layer LSTM networks, it can capture even more complex patterns. This paper explores the effectiveness of applying stacked LSTM networks in time series prediction domain, specifically, the traffic volume forecasting. Being able to predict traffic volume more accurately can result in better planning, thus greatly reduce the operation cost and improve overall efficiency. 
\end{abstract}


\section{Introduction}
With recent advancements in deep learning and the availability of huge amount of data,  data-driven prediction in time series has attracted more and more attention. Specifically, traffic forecasting is the key component of a transportation system powered by artificial intelligence\cite{duTrafficFlowForecasting2017} \cite{laiModelingLongShortTerm2017}. 
Traditionally, time series forecasting includes methods such as K-nearest Neighbor (KNN), Support Vector Regression (SVR), etc \cite{liBriefOverviewMachine2018}. This paper proposes the stacked LSTM model to capture the complex temporal patterns. The main contributions include:
\begin{itemize}
	\item Analyzing real traffic volume data in Toronto downtown area
	\item Proposing the stacked LSTM model with substantial gain comparing to the baseline model
	\item Improving the model performance with a comprehensive set of training methodologies
\end{itemize}

Deep learning has been widely adopted in time series forecasting\cite{laiModelingLongShortTerm2017}. One category is time series classification which assigns pre-defined class labels to time series output. The other is regression problem which predicts the actual value of unknown time series. There are efforts in modelling both long-term and short-term dependency patterns. This paper mainly focuses on the short-term dependency patterns.

In \cite{lvTrafficFlowPrediction2014}, a deep learning framework based on stacked auto-encoder is proposed to learn the representation of traffic flow features. It is trained in a greedy layer-wise fashion. In \cite{polsonDeepLearningShortterm2017}, it uses deep neural networks to capture these nonlinear spatial-temporal effects. The spatial and temporal dependencies are further discussed in \cite{yaoRevisitingSpatialTemporalSimilarity2019}. It argues that the spatial dependencies are dynamic, and temporal dependencies are not strictly periodic but with some perturbation. In addition to LSTM, some other neural networks such as the Convolutional Neural Networks (CNN) are also used in \cite{duTrafficFlowForecasting2017}.

\section{Problem Formulation}
Given a series of fully observed time series $\boldsymbol{X}=\left\{\boldsymbol{x}_{1}, \boldsymbol{x}_{2}, \ldots, \boldsymbol{x}_{T}\right\}$ where $\boldsymbol{x}_{t} \in \mathbb{R}^{n}$, our task is to predict the value $\boldsymbol{x}_{T+1}$ at the next time step. This is a regression problem. We can use a deep neural network model to predict $\boldsymbol{x}_{T+1}$ solely based on  $\boldsymbol{X}$ without considering any information further back in historical time steps. This is the baseline feedforward model with multiple layers. 

However, the baseline has its limitation in modelling the complex non-linear relationships between time steps. By leveraging the LSTM networks \cite{hochreiterLongShortTermMemory1997}, it can discover important temporal dependencies. The backpropagation through time in LSTM networks (more generally, recurrent neural networks) has vanishing and exploding gradients issues which limits the number of time steps it can learn. On the other hand, stacking multiple layers of LSTM will cause vanishing and exploding gradients issues as well. To improve model performance, this paper have adopted various techniques such as dropout layer \cite{srivastavaDropoutSimpleWay2014}, gradient clipping, early stopping, and adaptive learning rate, etc. 

\section{Model Architecture}
Traditional recurrent neural networks (RNN) \cite{stanfordcs231} is shown in Figure \ref{fig:rnn}, where $h_{t}$ denotes the hidden state at time step $t$, $x_{t}$ denotes the input at time step $t$, and $W$ denotes the parameter of neural networks. The relationship can be expressed as follows:
\begin{figure}
	\centering
	\includegraphics[width=0.6\textwidth]{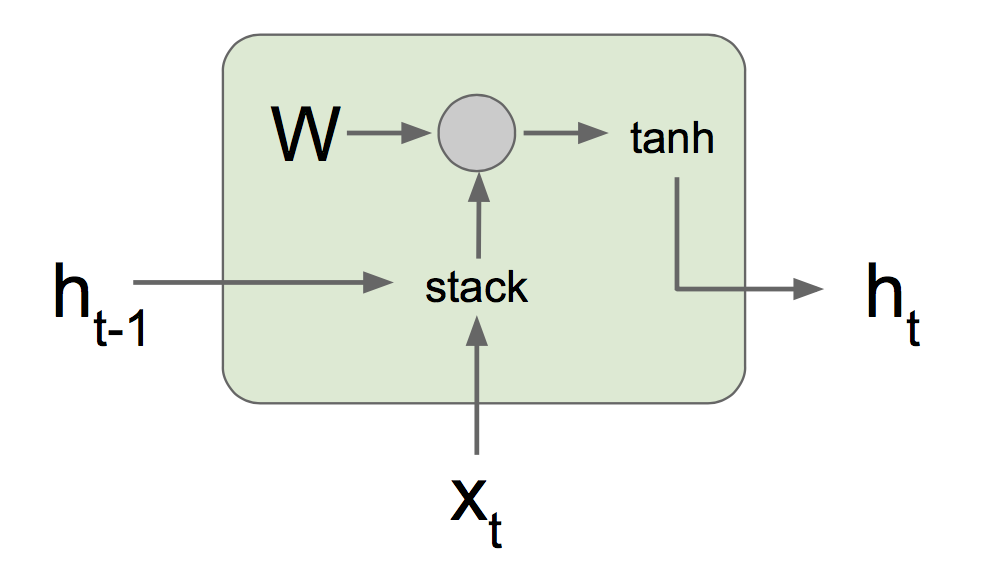}
	\caption{Traditional Recurrent Neural Networks.}
	\label{fig:rnn}
\end{figure}

\begin{equation}
\begin{aligned}
h_{t} &=\tanh \left(W_{h h} h_{t-1}+W_{x h} x_{t}\right) \\
&=\tanh \left(\begin{array}{cc}
\left.\left(W_{h h} \quad W_{h x}\right)\left(\begin{array}{c}
h_{t-1} \\
x_{t}
\end{array}\right)\right)
& =\tanh \left(W\left(\begin{array}{c}
h_{t-1} \\
x_{t}
\end{array}\right)\right)
\end{array}\right.
\end{aligned}
\end{equation}

However, it suffers from severe vanishing and exploding gradients issues, as the backpropagation involves repeated multiplication of matrix $W$ and $tanh$. If the largest singular value is $> 1$, then it has exploding gradients issue; otherwise if the largest singular value is $< 1$, then it has vanishing gradients issue.  

\begin{figure}
	\centering
	\includegraphics[width=0.7\textwidth]{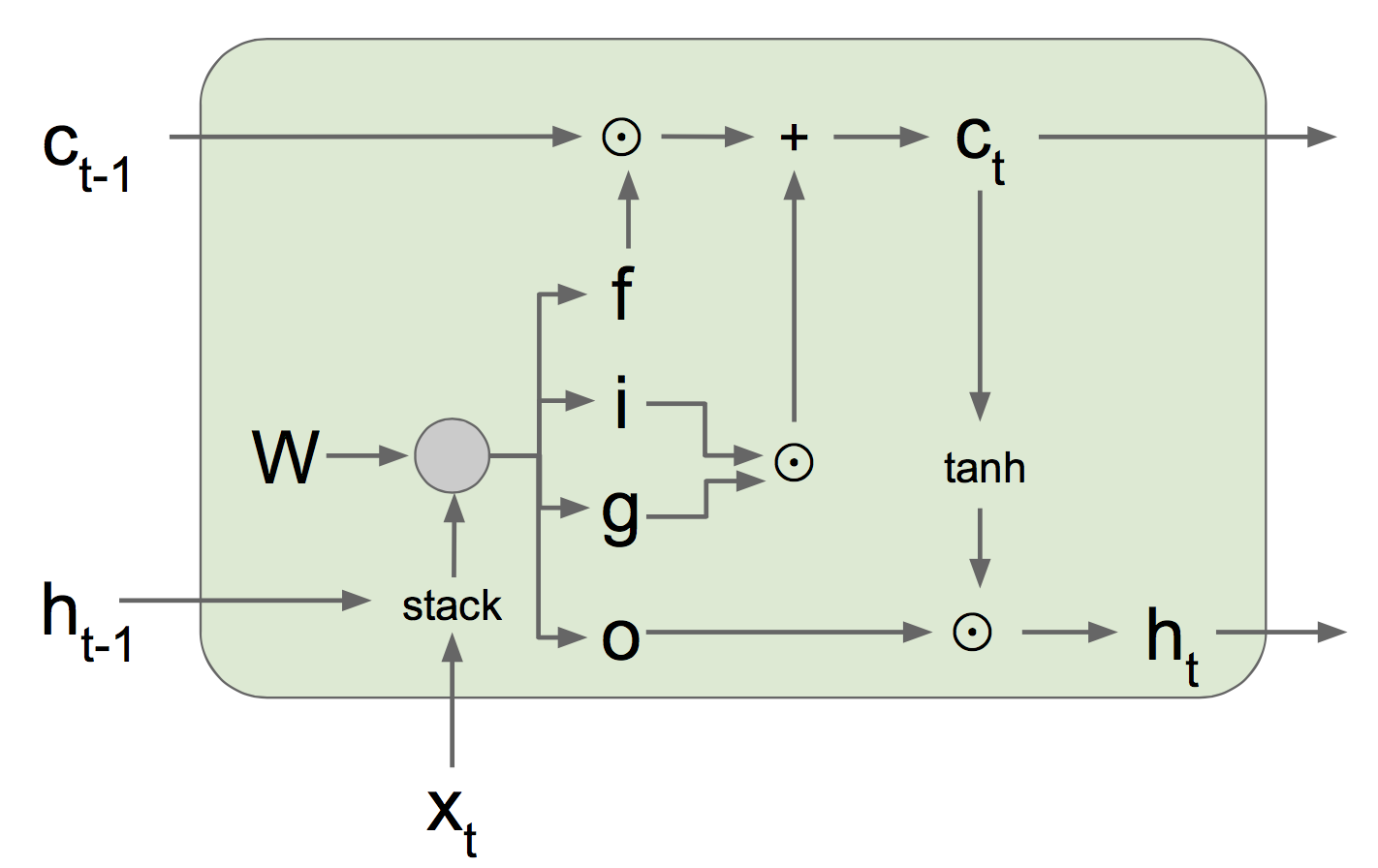}
	\caption{Long Short-Term Memory (LSTM) Networks.}
	\label{fig:lstm}
\end{figure}

To mitigate the vanishing and exploding gradients issues, LSTM networks introduce the cell state $c_{t}$, input gate $i$, forget gate $f$, and output gate $o$ as shown in Figure \ref{fig:lstm}. The relationship can be expressed as follows:

\begin{equation}
\begin{aligned}
\left(\begin{array}{l}
i \\
f \\
o \\
g
\end{array}\right) &=\left(\begin{array}{c}
\sigma \\
\sigma \\
\sigma \\
\tanh
\end{array}\right) W\left(\begin{array}{c}
h_{t-1} \\
x_{t}
\end{array}\right) \\
c_{t} &=f \odot c_{t-1}+i \odot g \\
h_{t} &=o \odot \tanh \left(c_{t}\right)
\end{aligned}
\end{equation}
where the $\sigma$ denotes the sigmoid activation function. 

With the help of cell state $c_{t}$, the backpropagation will have elementwise multiplication by the gate $f$ instead of the matrix multiplication by $W$. This greatly reduces the vanishing and exploding gradients issues. 

The whole model consists of multiple-layer LSTM networks with the last layer as a dense layer with single unit. The dense layer has linear activation functions to produce a real value number. 

\section{Training Tips and Tricks}
Some of the most common training tips in this experiment are listed below.

\subsection{Dropout}
\begin{figure}
	\centering
	\includegraphics[width=0.7\textwidth]{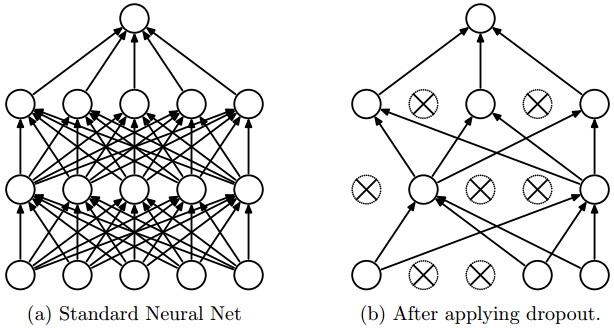}
	\caption{Dropout Mechanism.}
	\label{fig:dropout}
\end{figure}

Dropout is a regularization technique introduced by \cite{srivastavaDropoutSimpleWay2014} as shown in Figure \ref{fig:dropout}. During training, it sets a neuron output to zero with a certain probability. It can be considered as sampling a subnetwork within the full neural network, and only updating the parameters of the subnetwork. There are exponentially large number of subnetworks. During prediction, it keeps all neuron active. This can be considered as averaging predictions of those subnetworks. 

\subsection{Gradient Clipping}
This technique is to prevent the exploding gradients issue. One way to do gradient clipping is to clip the $L_{2}$ norm of the gradients when the norm exceeds the threshold. This has the advantage that each step is still in the gradient direction before clipping.

\subsection{Early Stopping}
\begin{figure}
	\centering
	\includegraphics[width=0.8\textwidth]{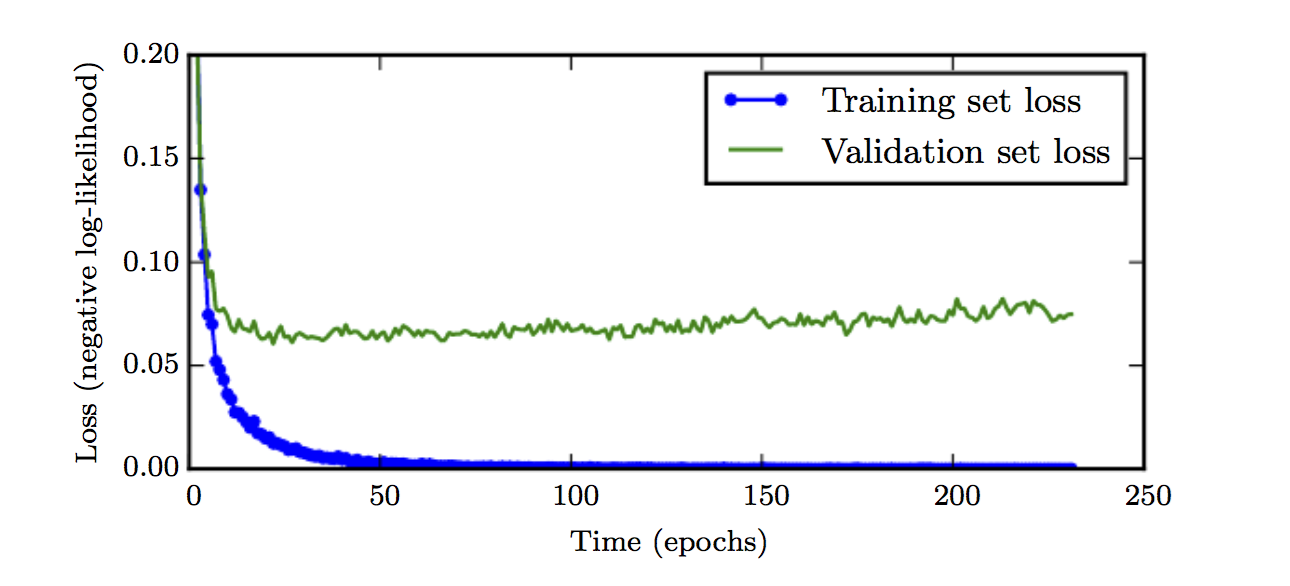}
	\caption{Training and Validation Loss over Epochs.}
	\label{fig:early_stop}
\end{figure}

The early stopping is a commonly used form of regularization in deep learning \cite{Goodfellow-et-al-2016}. The training and validation loss often behaves like Fig. \ref{fig:early_stop}. The training loss usually decreases over the iterations (i.e., epochs), however, the validation loss begins to increase. The model saved shall be the one with best generalization capability. Thus, the model with smallest validation error is saved.

\section{Experiments}
\subsection{Data}
\begin{figure}
	\centering
	\includegraphics[width=1.0\textwidth]{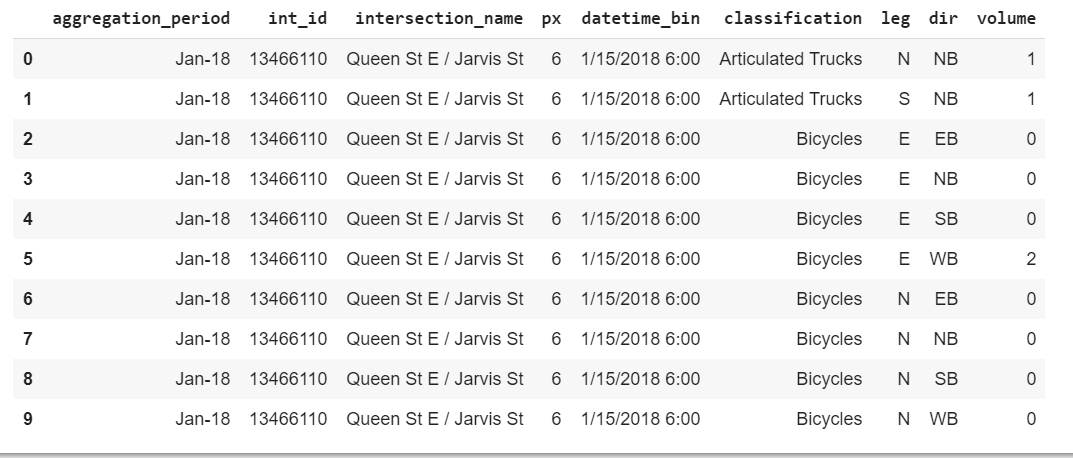}
	\caption{Source Data Format.}
	\label{fig:src_data}
\end{figure}

\begin{figure}
	\centering
        \includegraphics[width=0.7\textwidth]{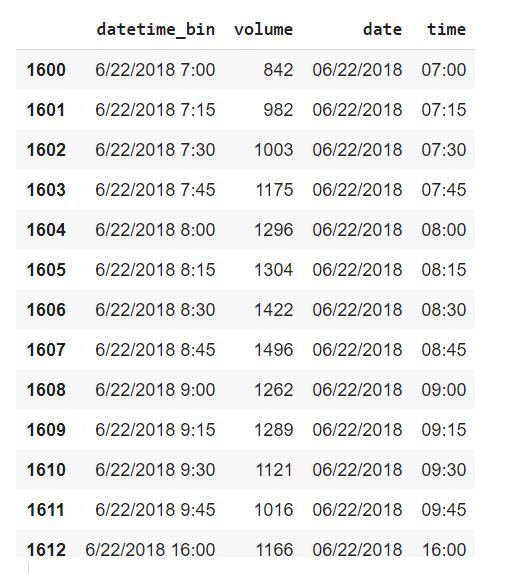}
	\caption{Processed Data Format.}
	\label{fig:processed_data}
\end{figure}

The original data format is shown in Fig. \ref{fig:src_data}. It records the intersection, date time bin, direction, type of vehicles and volumes. This experiment focuses on the total volumes of all vehicle types. The aggregated volume data is shown in Fig. \ref{fig:processed_data} with the time bin length of 15 minutes. 

The dataset is generated by a rolling window with length of 12 time bins. The next time bin will be considered as the label. Then, the whole dataset is split into train/validation/test dataset. Normalization is applied to the whole dataset.

\subsection{Results}

As a comparison, the baseline model has multiple dense layers without any long or short term memory mechanism.  The results are shown in Table~\ref{tab:table}. It contains Mean Absolute Error (MAE), Mean Square Error (MSE), and Root Mean Square Error (RMSE). The stacked LSTM model has much better performance than the baseline model, since it can capture longer temporal dependencies. 

\begin{figure}
	\centering
        \includegraphics[width=0.7\textwidth]{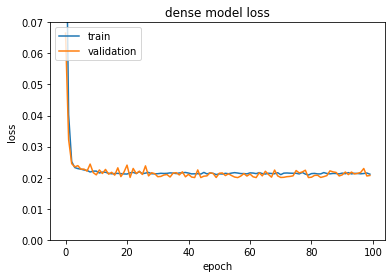}
	\caption{Training/Validation loss of Baseline Model.}
	\label{fig:dense_loss}
\end{figure}

\begin{figure}
	\centering
        \includegraphics[width=0.7\textwidth]{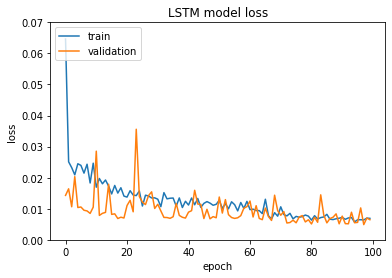}
	\caption{Training/Validation loss of Stacked LSTM Model.}
	\label{fig:lstm_loss}
\end{figure}

The training loss of baseline model and stacked LSTM are shown in Fig. \ref{fig:dense_loss} and Fig. \ref{fig:lstm_loss}, respectively. The baseline model has a much larger loss than the stacked LSTM model when the training converges. In addition, the training loss steadily decreases while the validation loss begins to increase. Thus, early stopping is used to select the best model with smallest validation loss. 

\begin{table}
	\caption{Evaluation Metrics}
	\centering
	\begin{tabular}{lll}
		\toprule
		& Baseline & LSTM \\
		\midrule
		MAE & 0.3976  & 0.1951     \\
		MSE     & 0.0902 & 0.0502      \\
		RMSE     & 0.3003       & 0.2241  \\
		\bottomrule
	\end{tabular}
	\label{tab:table}
\end{table}

\printbibliography

\end{document}